\documentclass{ieeeaccess}
\usepackage[frozencache=true,cachedir=minted-cache]{minted}

\usepackage{amsmath,amssymb,amsfonts}
\usepackage{algorithmic}
\usepackage{graphicx}
\usepackage{textcomp}
\def\BibTeX{{\rm B\kern-.05em{\sc i\kern-.025em b}\kern-.08em
    T\kern-.1667em\lower.7ex\hbox{E}\kern-.125emX}}

\usepackage[
sorting=none,
maxbibnames=99
]{biblatex}
\bibliography{main.bib}

\usepackage{comment}
\definecolor{codegreen}{rgb}{0,0.6,0}
\definecolor{codegray}{rgb}{0.5,0.5,0.5}
\definecolor{codepurple}{rgb}{0.58,0,0.82}
\definecolor{backcolour}{rgb}{0.95,0.95,0.92}
\usepackage{listings}

\lstdefinestyle{mystyle}{
    backgroundcolor=\color{backcolour},   
    commentstyle=\color{codegreen},
    keywordstyle=\color{magenta},
    numberstyle=\tiny\color{codegray},
    stringstyle=\color{codepurple},
    basicstyle=\ttfamily\footnotesize,
    breakatwhitespace=false,         
    breaklines=true,                 
    captionpos=b,                    
    keepspaces=true,                 
    numbers=left,                    
    numbersep=5pt,                  
    showspaces=false,                
    showstringspaces=false,
    showtabs=false,                  
    tabsize=2
}
\begin{document}
\history{Date of publication xxx, 2022, date of current version xxxx 00, 0000.}
\doi{10.1109/ACCESS.2017.DOI}

\title{Large Language Models are not Models of Natural Language: they are Corpus Models.}
\author{\uppercase{Csaba Veres}\authorrefmark{}}
\address[]{Department of Information Science and Media Studies, University of Bergen, Norway (e-mail: csaba.veres@uib.no)}


\tfootnote{This work was supported by the News Angler Project through the Norwegian Research Council under Project 275872.}

\markboth
{Veres \headeretal: Language Models are not Models of Language}
{Veres \headeretal: Language Models are not Models of Language}


\begin{abstract}
Natural Language Processing (NLP) has become one of the leading application areas in the current Artificial Intelligence boom. \emph{Transfer learning} has enabled large \emph{deep learning} neural networks trained on the \emph{language modeling} task to vastly improve performance in almost all downstream language tasks. Interestingly, when the language models are trained with data that includes software code, they demonstrate remarkable abilities in generating functioning computer code from natural language specifications. We argue that this creates a conundrum for the claim that \emph{eliminative} neural models are a radical restructuring in our understanding of cognition in that they eliminate the need for symbolic abstractions like \emph{generative phrase structure grammars}. Because the syntax of programming languages is by design determined by phrase structure grammars, neural models that produce syntactic code are apparently uninformative about the theoretical foundations of programming languages. The demonstration that neural models perform well on tasks that involve clearly symbolic systems, proves that they cannot be used as an argument that language and other cognitive systems are not symbolic. Finally, we argue as a corollary that the term \emph{language model} is misleading and propose the adoption of the working term \emph{corpus model} instead, which better reflects the genesis and contents of the model.
\end{abstract}


\titlepgskip=-15pt

\maketitle

\section{Introduction}

Deep-learning Artificial Neural Networks (ANNs) implement a multi-layered machine learning architecture which enables sophisticated representation learning \cite{nature14539}. They have significantly changed the technological, societal and commercial landscape in the past decade \cite{state,frontier}. In this article we focus on deep learning models for Natural Language Processing (NLP). \emph{Transformer} based deep learning models \cite{vaswani2017attention} have recorded significant improvements in various natural language tasks, and have entered service in industrial applications that have a significant linguistic component \cite{pandu,scott}. As an engineering artefact, these models have clearly enjoyed significant success. What is less clear is how much they have contributed to our understanding of natural language and cognitive architecture. While the most powerful systems have shown remarkable abilities complex language tasks like question answering and writing prose about arbitrary topics \cite{brown2020language}, there are contrary claims that ANNs are nothing but giant \emph{stochastic parrots}, with many hidden dangers if mis-represented as systems that \emph{understand} language in any non trivial sense \cite{bender2020/v1/2020.acl-main.463,bender}.

\emph{Language models} are joint probability distributions over sequences of words, or alternatively, functions that return a probability measure over strings drawn from some vocabulary \cite{manning2008introduction,bengio2003}. Large Neural LMs learn probability functions for sequences of real valued, continuous vector representations of words rather than discrete lexical items. Continuous representations are effective at generalising across novel contexts, resulting in better performance across a range of tasks \cite{bengio2003}. The probability distribution is learned through a form of \emph{language modeling}, where the task is to "predict the next word given the previous words" \cite{manning1999} (p.191) in word strings drawn from a corpus.

Neural LMs stipulate a mental architecture that stands in stark contrast with a Classical symbolic view \cite{FODOR19883} dominant in Linguistics since at least the beginning of the 20\textsuperscript{th} century, when language scholars began to study the structural properties of languages \cite{newmeyer,lasnik_2013}. The rigorous study of language as a cognitive faculty was pioneered by Noam Chomsky who introduced \emph{Generative Phrase Structure Grammars} as the rule structures underlying linguistic \emph{competence} \cite{1056813,structures}. In subsequent years a number of important challenges and modifications emerged, both from within and outside the research program\footnote{Chomsky describes his early work as an attempt to create a theoretical apparatus which was rich enough to describe the data. But it was always understood that the initial machinery had to be wrong because such a rich, complex system couldn't meet the criterion for biological evolution. The subsequent years were spent by reducing the complexity of the theoretical machinery \url{https://youtu.be/pUWmTXkpHjE?t=3520}} \cite{ewa}. Nevertheless, in the midst of technical controversies and disagreements, researchers have all agreed that "... linguistic knowledge is couched in the form of rules and principles." \cite{pinker/prince} (p.74).

The recent success of deep learning ANNs on wide ranging linguistic tasks have given support to the claim that statistical models are preferable to generative phrase structure grammars as theories of linguistic competence. Manning and Sch{\"u}tze argue in their classic text "Foundations of Statistical Natural Language Processing" that cognition in general and language in particular are "best formalized as probabilistic processes" \cite{manning1999} (p.15).  A similar position is more vigorously expressed by Peter Norvig in the context of a debate with Noam Chomsky at MIT\footnote{MIT150: Brains, Minds and Machines Symposium, June 16, 2011. Transcript available at \url{https://chomsky.info/20110616/}}: "Many phenomena in science are stochastic, and the simplest model of them is a probabilistic model; I believe language is such a phenomenon and therefore that probabilistic models are our best tool for representing facts about language" \cite{norvig}. Perhaps most dramatically, Geoff Hinton has proclaimed in his Turing Award acceptance speech that the success of "machine translation was the final nail in the coffin of symbolic AI" \cite{hinton/youtube} (32'30'').

In this paper we will defend the Classical symbolic view with an argument analogous to an indirect proof in logic, where the assumed truth of a premise leads to an absurd conclusion, thereby proving the falsehood of that premise. We present the argument as follows. In section 2 we provide some background by briefly explaining what is meant by a rule based, generative phrase structure grammar in natural language and software code. The argument itself begins in section 3 which details why ANN models are an alternative to explicit rule based grammar. The assumed premise is that the success of neural models makes them good explanatory models of natural language without the need for symbols. In section 4 we describe recent successes using ANN models - especially \emph{language models} - to automatically generate software code. But this is where the contradiction arises. If ANN models can be construed as explanatory theories for natural language based on their successes on language tasks then, in the absence of counter arguments, they should be good explanatory theories for computer language as well. We see no such argument and therefore arrive at the absurd conclusion that ANNs are good explanatory models of software. We know this is absurd because it is just a fact that the acceptable syntax of computer languages is determined by their grammar. Therefore, successful ANN models of natural language cannot be used as evidence \emph{against} generative phrase structure grammars in natural language. In section 5 we show that in fact \emph{language models} are most accurately described as \emph{corpus models}. The paper then concludes.

\section{Generative Phrase Structure Grammar for Natural Language and Software Code}

Linguistics in the first half of the 20\textsuperscript{th} century was mainly a taxonomic science with researchers pursuing \emph{Immediate Constituent (IC)} analysis inspired by Bloomfield \cite{Bloomfield}, and Saussure's \emph{structural linguistics} as outlined in his 1916 book, \emph{Course in General Linguistics}. The goal of early linguists was to develop methods that divide an expression into its immediate constituents, and continue the subdivision until syntactically indivisible parts were obtained. The essential insight contributed by Chomsky was that the constituent structure of language is the product of a system of \emph{rewrite rules} of the form $A \rightarrow \omega $ where $A$ is a class label and $\omega$ is a string that could contain \emph{terminal} strings as well as other class labels \cite{lasnik_2013/blevins}. An early example in \cite{structures} is the following simple grammar. (Note that the early formulations of grammar did not yet include the necessary machinery for recursive definitions \cite{lasnik_2013}.)
\begin{align}
&Sentence \rightarrow NP + VP\\
&NP \rightarrow T + N\\
&VP \rightarrow Verb + NP\\
&T \rightarrow the\\
&N \rightarrow man, ball, etc.\\
&Verb \rightarrow hit, took, etc.
\end{align}

This grammar can generate sentences of the type shown in Figure 1, through a series of \emph{derivations}.

\begin{figure}[h]
    \centering
    \includegraphics[scale=0.15]{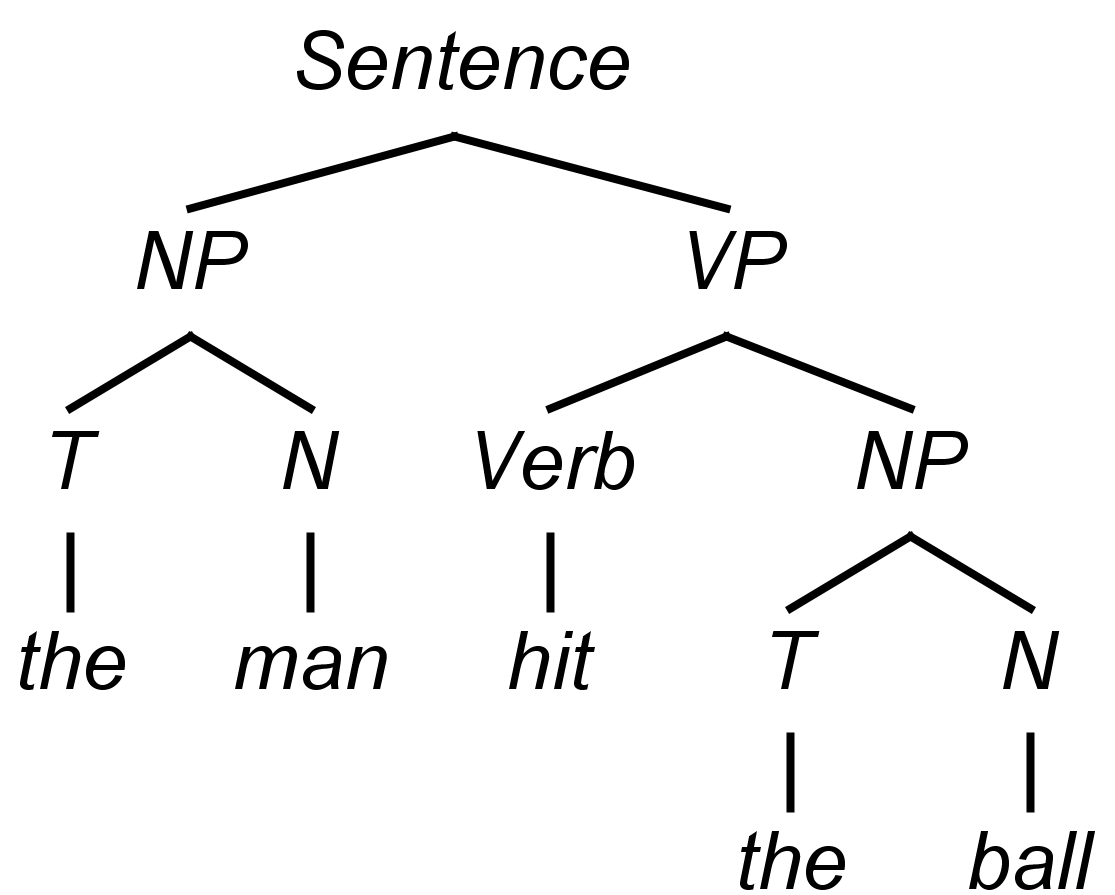}
    \caption{A tree diagram showing a result of a sequence of derivations using rules 1 - 6. It does not show the order of the derivation.}
    \label{fig:tree}
\end{figure}

\emph{Syntactic Structures} \cite{structures} outlined a new agenda for a formal linguistic theory in which \emph{language} was considered as " ... a set (finite or infinite) of sentences, each finite in length and constructed out of a finite set of elements" \cite{structures}(p. 13). Sentences are the output of a \emph{generative grammar}, and the Linguistics project is to explore grammars which generate all and only the \emph{grammatical} sentences of natural languages.

An important distinction was drawn between linguistic \emph{competence}, the speaker-hearer's knowledge of language, and \emph{performance}, actual instances of language use in concrete situations \cite{aspects}. A grammar on this view is a description of competence. It is not only a theory about possible sentence structure but about the intrinsic knowledge that allows the speaker to \emph{generate} an infinite set of grammatical sentences with a finite set of rules, and for the hearer to assign linguistic structure to each of those sentences. \emph{Performance}, on the other hand, includes the many psychological processes underlying actual linguistic productions, including effects of attention, memory, and so on. The multitude of these psychological processes are often little understood and can result in productions rife with errors from interference by a range of unpredictable, non language specific processes. For this reason Chomsky advocated using intuitions about grammatical acceptability as the primary data for linguistic theories rather than speech or text corpora.

One of the pioneers of high level computer programming languages, John W. Backus who led the Applied Science Division of IBM's Programming Research Group\footnote{\url{https://betanews.com/2007/03/20/john-w-backus-1924-2007/}} took inspiration from Chomsky's work on PSGs and conceived a \emph{meta-language} that could describe the syntax of languages that was easier for programmers to write than the machine languages of the time. The meta language later became known as \emph{Backus-Naur} form (BNF), so called partly because it was originally co-developed by Peter Naur in a 1963 IBM report on the ALGOL 60 programming language"\footnote{\url{https://www.masswerk.at/algol60/report.htm}}. 
The BNF is a notation for context free grammars consisting of \emph{productions} over \emph{terminal} and \emph{nonterminal} symbols, used to provide the precise syntax of programming languages which is required for writing compilers and interpreters for the language \cite{aho}. The original syntax described in the technical report included meta characters as shown in (7)
\begin{align}
\texttt{<  >  ::=  |}
\end{align}

where sequences of characters enclosed in the brackets represent metalinguistic variables and the ::= and $\mid$ are metalinguistic connectives. Productions or \emph{rewrite rules} are expressed using this machinery. For example the productions in (8) and (9) generate strings such as (10) and (11)
\begin{align}
\texttt{<ab> ::= ( | [ | <ab> ( | <ab> <d>} \\
\texttt{<d>  ::= 0 | 1 | 2 | 3 | 4 | 5 | 6} \\
\texttt{[(((1(36(} \\
\texttt{(12345(} 
\end{align}

BNF productions can be used to construct phrase structure descriptions of program expressions. For example given a simple grammar that includes the following production,

\begin{verbatim}
<statement> ::= 
    if <expression> then <statement>
    | if <expression> then <statement> 
      else <statement> 
    | <other>
\end{verbatim}

one can generate statement (12) where \emph{E\textsubscript{i}} are expressions and \emph{S\textsubscript{i}} are statements
\begin{align}
    \texttt{if \emph{E\textsubscript{1}} then if \emph{E\textsubscript{2}} then \emph{S\textsubscript{1}} else \emph{S\textsubscript{2}}}
\end{align}
 In turn the statement can be parsed to return the tree in figure 2 (example from \cite{aho}, p. 211)

\begin{figure}[h]
    \centering
    \includegraphics[scale=0.22]{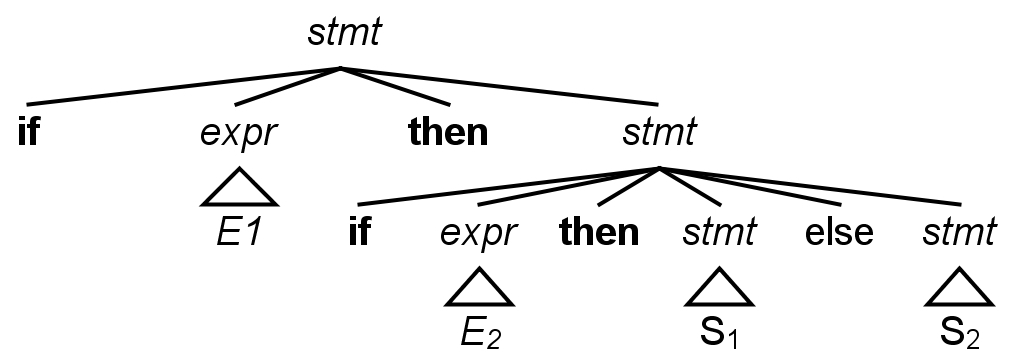}
    \caption{A parse of the statement if \emph{E\textsubscript{1}} then if \emph{E\textsubscript{2}} then \emph{S\textsubscript{1}} else \emph{S\textsubscript{2}}.}
    \label{fig:programtree}
\end{figure}

Chomsky's context free grammars were designed to handle the inherent ambiguities in natural languages, and the BNF inherited this property, which is undesirable for programming languages. One solution was proposed by Bryan Ford who introduced Parsing Expression Grammars (PEGs) which eliminated ambiguity with several new devices including \emph{ordered choice} \cite{ford}. In fact the current specification for Python 3.9 uses a mixture of BNF and PEG\footnote{\url{https://docs.python.org/3.9/reference/grammar.html}}.

\section{Language Without Rules}

The story of modern ANNs can be traced back to the Logical Calculus of McCulloch and Pitts \cite{mcculloch1943logical} who showed that it was possible to model the behaviour of networks of neurons with a logical calculus. This inspired researchers to create networks of artificial neurons with complex computational properties, which gave rise to the current generation of neural networks, the Deep Learning Networks \cite{lecun/nature14539}, which can learn complex non linear transformations implicated in image recognition, language processing, and other domains. The promise is that such models can explain complex cognitive phenomena such as language understanding, without the need for abstract symbol manipulating machinery.

Churchland argued that cognitive behaviour can be reduced to brain states, vis-\`a-vis parallel neural computations. Citing the work of Pellionis \emph{et al.} on the Tensorial approach to the geometry of brain function \cite{Pellionisz1980TensorialAT}, she asks us to imagine that "... arrays of neurons are interpretable as executing vector-to-vector transformations because that is what they really \emph{are} doing $-$ the computational problems a nervous system has to solve are fundamentally geometrical problems" \cite{churchland}(p.418). That is, transformations of real valued tensors is just what it is to be a cognitive agent.

Pinker and Prince \cite{pinker/prince} described this idea as \emph{eliminative connectionism} which are neural systems where it is impossible to find a principled mapping between "... the components of a PDP\footnote{Parallel Distributed Processing} model and the steps or memory structures implicated by a symbol-processing theory ...". Computations are performed at a non symbolic level, and any corresponding symbolic descriptions would only serve as short-hand approximations that could be used, for example, to make intuitive predictions \cite{pinker/prince}. The authors further distinguished eliminative connectionism from \emph{implementational connectionism} which is a class of systems in which the computations carried out by collections of neurons are isomorphic to the structures and symbol manipulations of a symbolic system. For example, \emph{recurrent neural networks} with long short-term memory have been shown to learn very simple context free and context sensitive languages \cite{gers}. More specifically the language with sentences of the form $a\textsuperscript{n}b\textsuperscript{n}$ is learned through \emph{gate units} acting as counters that can keep track of the number of terminal symbols in simple sequences \cite{gers}. It is therefore crucial to establish which kind of system deep learning models are, because eliminative neural systems are the only ones that offer a theoretical alternative to traditional grammar. Implementational systems would be fully compatible with a rules based linguistic theory and would therefore be theoretically uninteresting. 

According to Marcus \cite{marcus/cogp.1998.0694}, a clear criterion for recognising a genuinely eliminative connectionist system is how it implements \emph{compositionality}. Compositionality is a characteristic of Classical symbolic architectures, in which the representation-bearing computational units are structured expressions, and the operations performed on the expressions depends on their structure \cite{FODOR19883}. For example, $P$ follows from ${P\&Q}$ because of an operation which applies to the constituent structure of the representation. The $P$ and the $Q$ in the formula can be of arbitrary complexity and the operation will apply in exactly the same way. Eliminative ANNs are not Classical architectures because the operation that is responsible for transforming the network state from one which represents $P\&Q$ to one which represents $P$ does \textbf{not} operate by virtue of the constituent structure of the formula. Instead, the computation involves the network settling on an activation state representing $P$ following the presentation of $P\&Q$ because of the values of the model parameters which were adjusted to match the statistical association between $P\&Q$ and $P$ in the training data. In addition, if we substituted complex novel formulas for $P$ and $Q$ in a Classical system then the inference to $P\&Q$ would still hold, but the same would not be true in a network model that did not include the formulas in the training set. Finally if both $P$ and $Q$ were true in a Classical system then $P\&Q$ would also be true. On the other hand an ANN which was trained to output $P$ when $P\&Q$ was input would not output $P\&Q$ when presented with $P$ and $Q$ on the input, unless it was specifically trained to do so. 

There are good reasons to believe that deep learning networks are to be understood as non compositional, eliminative connectionist models. Bengio \emph{et al.} \cite{hinton/turing} argue that it is the non-linear transformations between the vectors in a deep learning architecture that allows the network to perform its functions, setting them apart from symbolic systems. As an example, "If Tuesday and Thursday are represented by very similar vectors, they will have very similar causal effects on other vectors of neural activity." \cite{hinton/turing} p.59. In a Classical system there is no inherent similarity between the two symbols "Tuesday" and "Thursday". Overlapping patterns of inference are only possible by asserting explicit axioms in the system to establish that the concepts are members of a collection that behave equivalently in certain circumstances. In a deep learning network, by contrast, the inferences are licensed by the similarity between the vectors themselves, which emerges through the learning process. "Tuesday" and "Thursday", on this account, have similar effects because they have overlapping representations learned from the sentences in the training corpus and not because they share common relations in an axiomatic system.

Additionally, Yun \emph{et al.} \cite{yun2020transformers} present a proof that deep learning transformers are universal approximators of continuous sequence-to-sequence functions\footnote{An additional condition is that the function must be \emph{permutation equivariant}, unless positional encodings are used.} with compact support, but a critical assumption of the proof is \emph{contextual mapping} of words in a sentence, such that the representation of individual words depends on the whole sentence. The proof requires, for example, that the "I" must be mapped to different vector embeddings in "I am happy" and "I am Bob". The semantic contribution of the constituent to the whole is a function of the whole, a direct contradiction to compositionality.

\subsection{Large Neural Language Models}

The most successful neural models for NLP are very large deep learning models that are trained on some version of the language modeling task, using vast amounts of text \cite{bommasani2021opportunities}. The models capture intricate statistical generalisations available in the training corpus which can subsequently be exploited in task specific training with much smaller data sets, a paradigm called \emph{transfer learning} \cite{bommasani2021opportunities}. Large LMs also appear to encode structural relations in language and can be probed in various ways to display aspects of their linguistic competence. For example Hewitt \& Manning \cite{hewitt-manning-2019-structural} argue that vector based representations in deep learning neural networks can capture syntactic tree dependencies between words without an explicit mechanism for constructing parse trees. Using structural probes with a learned, linear transformation, they argue that syntax trees are embedded "implicitly" in the deep learning model's vector geometry. Furthermore, they suggest that subject-verb agreement can be solved by using L2 distance metrics, since the verb that has to agree in number is closer to the subject than it is to any of the irrelevant attractor nouns following the transformation.

Goldberg \cite{goldberg2019assessing} reports a similar finding, that the BERT \cite{devlin2019bert} model with no additional fine tuning does remarkably well on predicting the subject appropriate number marking on the focus verb in sentences, even if there are mismatching distractors between the subject and the verb. He concludes that "exploring the \emph{extent} to which deep purely attention based architectures such as BERT are capable of capturing hierarchy-sensitive and syntactic dependencies — as well as the \emph{mechanisms} by which this is achieved — is a fascinating area for future research". In fact a sub field of machine learning that has affectionately come to be known as \emph{Bertology} \cite{rogers2020primer}, is attempting to do just that. 

In summary, neural LMs display some of the representations that are found in symbolic rule based theories, but it is not clear how they emerge or how they participate in computation. A potential explanation is suggested by Nefdt \cite{nefdt/s11023-020-09519-6} who proposed a refinement to the idea of compositionality with a distinction between \emph{Process-}, \emph{State-}, and \emph{Output-} \emph{Compositionality}. A system is \emph{process compositional} if the procedures it computes have \emph{meaningful parts}, and those parts are in principle \emph{knowable}. The meaningful parts are composed procedurally to give increasingly complex meaningful expressions. This is essentially the sense of compositionality we have ascribed to Classical systems. \emph{State compositionality} on the other hand describes representations in the system that can be \emph{decomposed} into smaller meaningful units even though they were not generated by a known compositional process. For example there are many analyses for ways in which lexical items might be decomposed into meaningful constituents (e.g. \emph{bachelor} = \emph{unmarried} + \emph{man}), but the decomposition does not exhaust the range of potential inferences the lexical item can enter into, nor is there a compositional process to explain how the constituents can be combined in learning the concept \cite{Fodor1981-FODTPS}. \emph{Output compositionality} takes a functional view in which for any input tokens \emph{t} there is a function \emph{f} which produces an output \emph{o} that is state compositional. Output compositionality is a special case of state compositionality because it does not imply that the entire analysis needs to be state compositional, it is sufficient that local outputs are. Neural language models would on this account exhibit output compositionality, such that various structured states can be observed, but these states do not reveal meaningful constituents which are causally responsible for the evolution of the system. The discovery of local compositional structure does not reveal how those structures participate in the evolution of the system (\cite{bengio2009learning,tenney,hinton/turing}).

\section{Neural Network Models and Software Code}

Hindle \emph{et al.} \cite{10.1145/2902362} proposed that the majority of software that people write can be described as \emph{natural programs} which are relatively simple, repetitive code that is used for communication with other programmers as much as it is to provide machine instructions. They successfully trained \emph{N}-gram \emph{language models} on software code to show that programming languages share many of the regularities observed in natural languages. They then used the trained model as a code suggestion plugin for the Eclipse Interactive Development Environment, resulting in keystroke savings of up to 61\% over the built in suggestion engine \cite{10.1145/2902362}.

Subsequent research expanded on this work by training significantly larger language models for mining source code repositories and providing new complexity metrics \cite{allamanis}, and using deep learning networks for code completion \cite{white}.

In a popular blog post, Andrej Karpathy drew our attention to the \emph{Unreasonable Effectiveness of Recurrent Neural Networks}\footnote{\url{http://karpathy.github.io/2015/05/21/rnn-effectiveness/} The title is a play on \emph{The Unreasonable Effectiveness of Mathematics in the Natural Sciences}, a 1960 article by the Nobel Prize winning physicist Eugene Wigner.} with some illustrations of RNNs learning from character level data. He showed that models could learn to generate natural language prose, prose which includes markdown, valid XML, LaTeX, and source code in the C programming language. The RNN generated both LaTeX and C code with remarkable syntactic fidelity including spacing, matching brackets, variable declarations and so on. Observed errors were principally semantic. For example, one example of LaTeX code had a 13 line environment beginning with \verb|\begin{proof}| and ending with \verb|\end{lemma}|. That is, the coreference dependency is partially broken; while the \verb|\begin{}| block is closed as required by the syntax, the RNN has not correctly remembered the exact identity of the block. Similarly the C code contains very few syntactic errors, but contains semantic errors such as variable names not being used consistently, or using undefined variables.

\subsection{Large Neural Language Models and Code Synthesis}

Hendrycks \emph{et al.} \cite{hendrycksapps2021} released APPS, an ambitious dataset and benchmark for measuring the effectiveness of machine learning models in a realistic code generation framework, involving natural language problem specifications and functional test cases. The problems have three levels of difficulty; introductory, interview, and competitive. The following example is the natural language description in an "interview" level problem. 

\begin{quote}

You are  given two integers \emph{n} and \emph{m}. Calculate the number of pairs of arrays (\emph{a}, \emph{b}) such that: the length of both arrays is equal to \emph{m}; each element of each array is an integer between 1 and \emph{n} (inclusive); a\textsubscript{i} $\leq$ b\textsubscript{i} for any index \emph{i} from 1 to \emph{m}; array \emph{a} is sorted in non-descending order; array \emph{b} is sorted in non-ascending order. As the result can be very large, you should print it modulo 10\textsuperscript{9} + 7.

Input: The only line contains two integers \emph{n} and \emph{m} (1 $\leq$ \emph{n} $\leq$ 1000, 1 $\leq$ \emph{m} $\leq$ 10). Output: Print one integer - the number of arrays \emph{a} and \emph{b} satisfying the conditions described above modulo 10\textsuperscript{9} + 7.

-----Examples-----

Input: 2 2,  Output: 5

Input: 10 1, Output: 55

Input: 723 9, Output: 157557417

\end{quote}

The authors gathered 10,000 coding problems with 131,836 test cases for checking the generated solutions, and 232,444 gold-standard solutions written by humans. They tested four models for their ability to generate code that could solve the test cases; code which is both syntactically well formed and semantically correct. The models they tested were GPT-2 \cite{Radford2019LanguageMA} (two versions, with 0.1 and 1.5 Billion parameters), GPT-Neo (2.7B parameters) \cite{gpt-neo} and GPT-3 \cite{brown2020language} (175B parameters). The GPT-2 models received additional pre-training with 30 GB of Python code from GitHub. GPT-Neo was trained on the Pile \cite{gao2020pile} which already contains code from GitHub. GPT-3 had no prior training on code. All but GPT-3 were then further fine tuned on the APPS dataset, while GPT-3 was tested in a few-shot scenario. 

The results showed that in general bigger models were better than the smaller models, and pre-training on code is critical since the largest GPT-3 model without pre-training (or fine tuning) was outperformed by the smallest GPT-2 model. 

In general this proved to be a difficult task and the best model passed only 14.68\% of test cases on the introductory problems, and 6.54\% for the 'competitive' problems. On a strict criterion where success was defined as passing all of the test cases, these figures dropped to 3.9\% and 0\% respectively. However, the important finding for our argument was the extremely low rate of syntax errors in the code. Remarkably, GPT-Neo generated code that failed because of syntax errors on only 3\% of its solutions on the "introductory" problems, increasingly to approximately 5\% for the "competition" problems.

Listing 1 shows a machine synthesized solution for the the example problem. The solution attempts to build arrays as described, to return a result which depends on those arrays, and to provide the answer as modulo 10\textsuperscript{9} + 7. However the logic it uses to build and compare the arrays is faulty and the results incorrect.

\begin{listing}[!ht]
\begin{lstlisting}[language=Python, caption=Example solution from a model tested on APPS. This solution passes 0 test cases but it is a legal Python program.] 

               
n, m = map(int, input().split()) 
mod = 10**9+7 

fact = [1, 1] 
factinv = [1, 1] 
inv = [0, 1] 

for i in range(2, n+1): 
    fact.append((fact[-1] * i) % mod) 
    inv.append((-inv[mod % i] * (mod//i)) % mod) 
    factinv.append((factinv[-1] * inv[-1]) % mod) 
    
def cmb(n, r, mod): 
    if (r < 0) or (n < r): 
        return 0 
    r = min(r, n - r) 
    return fact[n] * factinv[r] * factinv[n-r] % mod 
    
print((cmb(n, m, mod) * cmb(m-1, n-1, mod)) % mod)
\end{lstlisting}
\label{listing:1}
\end{listing}

Austin \emph{et al.} \cite{austin} released a somewhat easier dataset, the Mostly Basic Programming Problems (MBPP), comprising natural language descriptions, concrete test cases, and gold standard solutions. They argued that the poor model performance reported by \cite{hendrycksapps2021} was due to the difficult nature of the problems. In particular the competition set was written in a manner that purposely obfuscated the connection between problem statement and solution code. The MBPP was therefore designed to maintain the clarity of the natural language descriptions and to be solvable by entry-level programmers. The problems and solutions were provided by crowd-sourced workers and subsequently inspected by the authors. The models used for the experiments were BERT-style transformer models \cite{vaswani2017attention} with parameter counts ranging from 244 million to 137 billion, and pre-trained on a combination of web documents, dialog data, and Wikipedia. The pre-training dataset contained 2.97B documents, of which 13.8M were web sites that contained some code and text, although complete source code files were not included. 

The experiments showed a log-linear improvement in performance with model size, and only marginal improvement of fine tuning over the pre-trained models. The largest model with 137 billion parameters could solve approximately 60\% of the problems with at least one generated solution, while the smallest 244 million parameter model could solve fewer than 5\%. The models benefited from fine-tuning by approximately a constant 10\% from the 422 million to the 68 billion parameter model. As a result the benefit of fine-tuning decreased proportionately as model size increased. 

The qualitative error analysis revealed that the main reason for failure was not syntactic errors but problems with the code semantics. Even the smallest models produced syntactically correct Python code about 80\% of the time, increasing to over 90\% for the larger models. The most commonly observed semantic errors were in complex problems with multiple constraints or sub-problems, where the generated code only solved one sub problem. For example the problem \emph{"Write a function to find the longest palindromic subsequence in the given string"} might result in code that found just one of the palindromic sequences but not the longest one. Another possible mistake was code that solved a problem that was similar, but more common than the provided one. For example the query \emph{“Write a python function to find the largest number that can be formed with the given list of digits."} might result in a program that calculated the largest digit in the list. 

One difficulty with this finding is that it is not possible to identify the exact locus of the semantic errors. It could, for example be the natural language component which fails to extract the correct specification. In order to investigate the level of semantic understanding in the code synthesis component, the authors tested to see how well the model could predict the output of the ground truth source code. If the model had some understanding of the semantics of code then this should manifest itself in learning code execution. Previous work has found that different architectures were in fact capable of learning to execute code \cite{NEURIPS2020_62326dc7}. In a few-shot setting where only the code was presented as a prompt and the model had to predict the output in an input-output test case, the accuracy was 16.4\%, and improved to 28.8\% when two input-output example pairs were also presented. These results dropped to 8.6\% and 11.6\% respectively when 2 test cases were presented (and the model had to answer both correctly). More interestingly, when only the natural language problem description and an example input-output prompt was presented the model performed more accurately, at 12.8\%. Finally, the model performed better when it was presented with only the examples as prompt then it did with the code itself, with no example (10.2\% example only, 8.6\% code only). Together the results suggest that the model was not considering the semantics of the code at all when predicting the results of execution. The model appeared to be using previously observed patterns rather any understanding of the code.

OpenAI Codex\footnote{https://openai.com/blog/openai-codex/} is an experimental API from the Microsoft owned company which powers GitHub Copilot\footnote{https://copilot.github.com/}. Codex is a GPT-3 based model fine tuned on publicly available code, which generates Python code from docstrings \cite{chen2021codex}. For testing purposes the authors introduced the HumanEval evaluation set which includes 164 manually constructed programming problems consisting of a function signature, docstring, body, and several unit tests. Listing 2 shows an example problem together with a correct and an incorrect solution. The incorrect solution supports Austin \emph{et al.}'s analysis that the errors are not syntactic but semantic, in this case caused by solving only a part of the problem. In this example the generated algorithm fails to consider the boundary condition where n = 1.

\begin{listing}[!ht]
\begin{lstlisting}[language=Python, caption=Example HumanEval docstring and a correct and incorrect completion]    

def is_prime(n): 
    """Return true if a given number is prime, and 
        false otherwise. 
    >>> is_prime(6) False 
    >>> is_prime(101) True 
    >>> is_prime(11) True 
    >>> is_prime(13441) True 
    >>> is_prime(61) True 
    >>> is_prime(4) False 
    >>> is_prime(1) False 
    """
    ### COMPLETION 6 (CORRECT): ###
    prime = True
    if n == 1:
        return False
    for i in range(2, n):
        if n % i == 0:
            prime = False
    return prime
        
    ### COMPLETION 7 (WRONG): ###
    for i in range(2, n):
        if n % i == 0:
            return False
    return True
    
\end{lstlisting}
\label{listing:2}
\end{listing}

They evaluated (a) a 12B parameter vanilla GPT-3 model, (b) Codex, which is GPT-3 fine-tuned on 159GB of code collected from GitHub, and (c) Codex-S which is further fine-tuned on correctly implemented standalone functions. When looking at single solutions the vanilla GPT-3 model was unable to solve any HumanEval problems, while the 12B parameter Codex solved 28.8\%. Fine-tuning this model on function solutions produced Codex-S which solved 37.7\% of the problems, which was a bigger gain from fine-tuning than observed on the other studies. When considering multiple solutions, Codex-S was able to find at least one correct solution 72.31\% of the time from among 100 candidate solutions, and 46.81\% of the time among 10 solutions. 

Probing error points once again provided some interesting insights. The authors created "building blocks" which were simple tasks such as “convert the string s to lowercase”, “remove all vowels from the string”, and “replace spaces with triple spaces”. When the building blocks were systematically concatenated to form increasingly complex function descriptions, model performance decreased exponentially with increasing problem complexity. This pattern is not likely to be observed with human programmers, as the task is simply to link simple discrete functions together. 

In summary, deep learning neural network models are beginning to have impressive success in generating syntactically correct code from natural language specifications. The various models demonstrate not only a knowledge of programming constructs like variable declaration and operation signatures, but also constructs that depend on constituency structure like long distance dependencies (e.g. matching brackets, matching if-then imperatives). On the other hand the knowledge of semantics is either absent or very limited. 

\section{Large Neural Language Models are Corpus Models}

The surprising success of neural LMs to synthesize computer code has resulted in some puzzling claims. For example \cite{austin} speculate that for models "it is not necessary to explicitly encode the grammar of the underlying language $-$ they learn it from data." Taken literally the claim would be that the neural network is implementational after all, since it is learning the grammar of a programming language. We have already pointed out the problems with this position when applied to models of natural language, and suggest that similar problems exist in the domain of programming languages.

The most parsimonious hypothesis is that language models perform natural language and programming tasks through the same mechanisms, which is not that they learn the grammar of the language. The evidence that neural network representations can encode hierarchical components in language  (e.g. \cite{hewitt-manning-2019-structural,tenney}) presents the possibility that the same representations are used in encoding structural relationships in software code. Some support for this comes from \cite{chrupala} who developed a novel method for probing neural LMs for their ability to encode natural language syntax. In order to validate their method they used neural models trained on a synthetic language for arithmetic expressions with a simple syntax, concluding that the neural network encoding of the structural patterns generated by an artificial grammar are analogous to their encoding of natural language structures. 

It is important to point out once again that programming languages have an exact grammar which is, by hypothesis, not learned by the LM. Therefore the LM's ability to abstract hierarchical relationships by some means other than a grammar, is a powerful tool which can result in impressive linguistic performance as well.

It is clear that large LMs are able to generate syntactically well formed productions in domains where they receive large amounts of training data, whether that be natural language or software code. Since the data is their only resource for learning, they can be susceptible to the characteristics of the data set. For example, Bender \emph{et al.} \cite{bender} documented many different selection biases which can affect which utterances are included in corpora. Perhaps even more problematical is that the performance of language models on language tasks is strongly influenced by the diversity of the training corpora \cite{gao2020pile,TuringNLG}, which shows that what is modeled is not language per se, but the characteristics of the training corpora. Similarly Brown \emph{et al.} \cite{brown2020language} used a filtered version of the Common Crawl dataset\footnote{https://commoncrawl.org/the-data/}, a collection of text based on 12 years of web crawling, and found that the unfiltered version gave poorer results on language tasks than the higher quality version which was filtered in part through a comparison with text in written books.

If the quality of a dataset can influence the performance of the model, then there should be clear guidelines as to what constitutes a high quality dataset. To the best of our knowledge the answer to this in the neural LM literature is rather vague. Take for example the Pile, a dataset specifically designed to be a large, diverse and high-quality resource for machine learning \cite{gao2020pile}. Unfortunately the authors are not explicit about what they regard as "high-quality", even though they use the descriptor 33 times in their paper. Some clues emerge through the description of individual data sources such as Wikipedia which is "... a source of high quality, clean English text ..." (p.4), suggesting as one criterion a text written in complete English sentences with some editorial control. Also mentioned is the Common Crawl web content where one challenge is "... difficulty of cleaning and filtering the Common Crawl data ..." due to low level problems such as presence of non text characters, segmentation problems, capitalization, etc. \cite{manning1999}. Common Crawl data also contains duplicate lines which often signals highly repetitive "boilerplate text", which can reduce performance on downstream tasks \cite{lee}. Grave \emph{et al.} \cite{grave} tried to filter Common Crawl by using Wikipedia "because the articles are curated, the corresponding text is of high quality". Again, editorial control is an important factor. Another technique for increasing quality by comparing against Wikipedia was used by Wenzek \emph{et al.} \cite{wenzek} who measured the perplexity score of target web pages against a language model trained on Wikipedia, and took the perplexity as a measure of quality. Again the strategy is to assume Wikipedia is high quality and use it to filter other content.

The practice is problematic in our view because it is important to be specific about what should be accepted as appropriate input to a "language model", as the data is literally the "language" in the model. In linguistic theory, as we saw earlier, Chomsky proposed grammatical acceptability judgement as a way to decide which utterances counted as valid data for linguistic enquiry. But this view is criticised in statistical NLP. For example Manning \cite{manning:2003} argued against the Chomskyan notion of \emph{grammaticality} and proposed probabilistic syntax largely on the strength of examples which purportedly show the non categorical nature of grammar. Norvig \cite{norvig} argued even more forcefully that "... people have to continually understand the uncertain, ambiguous, noisy speech of others ... Chomsky for some reason wants to avoid this, and therefore he must declare the actual facts of language use out of bounds and declare that true linguistics only exists in the mathematical realm ... Chomsky dislikes statistical models in that they tend to make linguistics an empirical science (a science about how people actually use language) ..." So for Norvig the data for the study of lnguage is the "actual use" of language which are the noisy and uncertain utterances of the everyday. But this view makes it difficult to justify eliminating any "low quality" utterances reflecting how people "actually use language" from the corpus. If we are going to judge dataset quality by comparing it against the edited and grammatical text in "high quality" sources such as Wikipedia, then we need to be careful with our theoretical assumptions. The problem is that we might be smuggling \emph{grammaticality} in by the back door.

For the preceding reasons we would suggest a clarification in terminology, and propose a change from the theory-laden term \emph{language model} to the more objectively accurate term \emph{corpus model}. Not only does the term \emph{corpus model} better reflect the contents of models, it also provides transparency in discussing issues such as model bias. One might be surprised if a \emph{language model} is biased, or if there is different bias in two different \emph{language models}, but a bias in \emph{corpus models} and different biases in different \emph{corpus models} is almost an expectation. Natural language is not biased. What people say or write can be biased.

\section{Conclusion}

We considered the challenge that deep learning neural models present to traditional generative phrase structure theories of natural language, and showed the challenge to be invalid since the arguments could equally and absurdly be applied against phrase structure in software code. We claim our argument to be more powerful and permanent than more traditional attempts to prove that neural networks \emph{cannot} perform particular tasks.  This style of argument has an illustrious history in the field, where research activity was drastically reduced for decades by the publication of Minsky and Pappert's \emph{Perseptrons}, which showed fundamental computational limitations of the contemporary neural models \cite{perceptron}. More sophisticated models were eventually devised to overcome these limitations, and the previously devastating criticisms vanished. Our argument is more powerful because it is predicated on the \emph{success} of neural models. We argue that achieving high performance on arbitrary NLP tasks is \textbf{irrelevant} to theories of natural language in general, and generative grammar in particular, because the same arguments can apply equally to programming languages which are clearly the product of a generative grammar.

As a corollary we argued that the term "language model" is misleading. A more accurate and useful term would be "corpus model". We hope this clarification is useful for practical work as well as scientific discovery, and look forward to theoretical insights that can be gained by exploring the similarities between the statistical epiphenomena created by natural languages and computer software in shared corpus models. 

Pinker and Prince argued that the connectionist models of the time failed to deliver a "radical restructuring of cognitive theory" \cite{pinker/prince}(p.78), because they did not adequately model relevant linguistic phenomena. We argue that modern neural models similarly fail, but from the opposite perspective. In becoming universal mimics that can imitate the behaviour of clearly rule driven processes, they become uninformative about the true nature of the phenomena they are "parroting" \cite{bender}. Enormous amounts of training data and advances in compute power have made the modern incarnation of artificial neural networks tremendously capable in solving certain problems that previously required human-like intelligence, but just like their predecessors, they have failed to deliver a revolution in our understanding of human cognition.

\printbibliography

\begin{IEEEbiography}[{\includegraphics[width=1in,height=1.25in,clip,keepaspectratio]{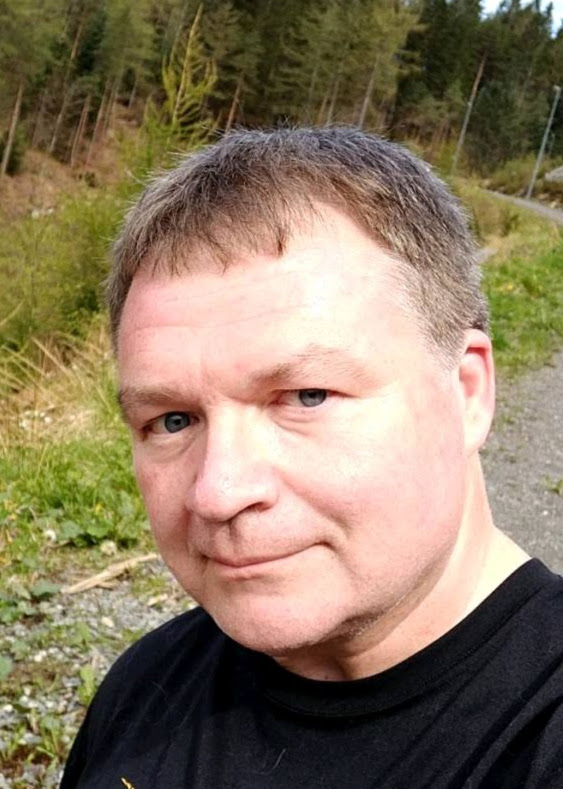}}]{Csaba Veres} was born in Budapest in 1964. Csaba received the Ph.D. degree in cognitive science from the University of Arizona, Tucson. His thesis was in the field of Psycholinguistics, where he studied the role of meaning on sentence parsing and representation. He subsequently began work as a computer and information scientist at Melbourne University in Australia. 

He is currently Full Professor at the Department of Information Science and Media Studies at the University of Bergen, Norway. His areas
of expertise include NLP, machine learning, and
semantic web technologies. He has experience as an Academic
and Practitioner. He founded a Norwegian company called LexiTags, and consulted as Head of AI with the London based educational technology startup, Flooved. He has published a wide assortment of original research articles, and has had popular linked data apps on the Apple store, called MapXplore and AuotoMind. He has also held positions as a Research Scientist at the Australian Defence Science and Technology Organisation, and as a Senior Lecturer in the Department of Information Systems, Melbourne University.

\end{IEEEbiography}

\EOD

\end{document}